%% file: neurips_2026.tex
\documentclass{article}

\usepackage[main, final]{neurips_2026}

\usepackage[utf8]{inputenc}
\usepackage[T1]{fontenc}
\usepackage{hyperref}
\usepackage{url}
\usepackage{booktabs}
\usepackage{amsmath}
\usepackage{amssymb}
\usepackage{amsfonts}
\usepackage{nicefrac}
\usepackage{microtype}
\usepackage{xcolor}
\usepackage{graphicx}
\usepackage{array}
\usepackage{multirow}

\title{CodecSplat: Ultra-Compact Latent Coding for Feed-Forward 3D Gaussian Splatting}

\author{%
  Pengpeng~Yu$^{1,2}$, Runqing~Jiang$^{1}$, Qi Zhang$^{2,3}$ \\
  \textbf{Dingquan~Li}$^{2}$, \textbf{Jing~Wang}$^{2\dagger}$, \textbf{Yulan~Guo}$^{1,2\dagger}$ \\
  $^{1}$Sun Yat-sen University, China\\
  $^{2}$Pengcheng Laboratory, China\\
  $^{3}$Peking University, China\\
  \texttt{yupp5@mail2.sysu.edu.cn}\\
}

\begin{document}
\maketitle

\renewcommand{\thefootnote}{\fnsymbol{footnote}}
\footnotetext[2]{Corresponding author.}
\renewcommand{\thefootnote}{\arabic{footnote}}

\begin{abstract}
While feed-forward 3D Gaussian splatting reconstructs renderable Gaussian primitives from sparse context views without per-scene optimization, existing pipelines do not provide a compact scene representation for storage or transmission. A natural solution is to apply existing 3DGS compression methods to the generated Gaussian primitives. However, this approach operates on the final irregular 3D representation and is decoupled from the internal feature-to-Gaussian generation process, which limits compression efficiency. To address this, we introduce \emph{CodecSplat}, an ultra-compact latent coding framework for feed-forward 3D Gaussian splatting. CodecSplat first encodes an intermediate 2D Gaussian-generation feature into an entropy-coded scene bitstream. At the decoder, the latent feature is reconstructed and used to predict depth and Gaussian parameters, which are then mapped to 3D Gaussian primitives. By integrating compression into the feed-forward Gaussian generation pipeline, CodecSplat avoids inefficient compression over irregular 3D Gaussian primitives and allows the codec to exploit the structured intermediate feature representation.
We instantiate CodecSplat on a feed-forward Gaussian splatting backbone with depth-guided multi-view feature refinement and a hierarchical learned feature codec.
On DL3DV and RealEstate10K datasets, CodecSplat achieves 23.56--26.36 dB and 24.76--27.05 dB PSNR with only 20.00--107.77 KiB and 3.37--12.51 KiB per scene, respectively.
This is roughly one order of magnitude smaller than compressing feed-forward generated Gaussian primitives, while preserving controllable rate--distortion behavior.
Code is available at \url{https://github.com/pengpeng-yu/CodecSplat}.
\end{abstract}

\section{Introduction}
\label{sec:introduction}

3D Gaussian Splatting (3DGS) represents a scene as a set of Gaussian primitives with learnable positions, covariances, opacities, and view-dependent colors, and renders them efficiently with differentiable rasterization~\citep{kerbl2023gaussians}.
Despite its high rendering quality, the original 3DGS formulation requires optimizing Gaussian parameters separately for each scene, which limits its scalability in applications that demand rapid reconstruction of many unseen scenes.

Feed-forward 3D Gaussian splatting (FF3DGS) addresses this limitation by predicting Gaussian primitives directly from sparse context images and camera parameters.
Numerous methods~\citep{charatan2024pixelsplat,chen2024mvsplat,xu2025depthsplat,xu2026resplat} learn generalizable mappings from posed images to renderable Gaussian representations.
These works generally combine multi-view matching, depth or geometry estimation, and pixel-aligned Gaussian prediction, thereby avoiding iterative scene-specific optimization and enabling fast sparse-view novel view synthesis.

Despite this progress, compact representations for feed-forward Gaussian scenes have received less attention.
Existing 3DGS compression methods are mostly developed in the per-scene optimized 3DGS setting, where the Gaussian representation and compression-related components are optimized for each scene through pruning, quantization, learned entropy modeling, or anchor-based context modeling~\citep{fan2024lightgaussian,navaneet2024compact3dgs,chen2024hac,wang2024contextgs,wang2024rdogaussian,lee2025codecgs}.
These methods are well matched to per-scene optimized Gaussian assets, but are suboptimal for FF3DGS outputs because they compress Gaussian primitives after a structured-to-irregular transformation, where spatial locality and cross-view redundancy have already been largely lost.
Unlike per-scene optimized 3DGS, FF3DGS generates Gaussian primitives through a feed-forward image-to-Gaussian pipeline.
A straightforward way to reuse existing compressors is to first execute the FF3DGS model and then compress the generated Gaussian primitives.
However, this approach still needs to encode a large set of irregular 3D Gaussian primitives, making the codec operate on a less structured representation.
More importantly, it treats the generated primitives as a standalone compression target and is decoupled from the internal feature-to-Gaussian generation process.
Before the final primitives are formed, the feed-forward network has typically organized scene information into dense, view-aligned, and geometry-aware features on regular 2D grids. Therefore, compressing only the generated primitives discards this structured intermediate representation and can degrade compression efficiency.
This observation motivates us to move the coding point earlier in the pipeline by encoding the intermediate representation before it is mapped to 3D Gaussian primitives.

To obtain a compact scene representation for FF3DGS, we propose \emph{CodecSplat}, a learned latent coding framework integrated into the feed-forward Gaussian generation pipeline.
On the encoder side, CodecSplat performs multi-view depth and feature extraction, refines the Gaussian-generation feature with depth-guided multi-view interaction, and entropy-codes the resulting latent feature into a scene-level bitstream.
On the decoder side, CodecSplat entropy-decodes the latent representation, predicts depth and Gaussian attributes with lightweight heads, and maps the pixel-aligned predictions to 3D space. 
By placing the coding bottleneck before 2D-to-3D Gaussian construction, CodecSplat allows compression and feed-forward Gaussian generation to operate on the same compact intermediate representation.
Experiments on DL3DV and RealEstate10K show that this design yields KB-level scene bitstreams while preserving controllable rate--distortion behavior.

Our contributions are summarized as follows:
\begin{itemize}
  \item We introduce an integrated compression paradigm for feed-forward 3D Gaussian splatting. By placing the coding bottleneck inside the feed-forward image-to-Gaussian pipeline, the proposed formulation directly produces a compact 3DGS representation and avoids a separate export-and-compress stage over irregular Gaussian primitives.

  \item We propose CodecSplat, a learned latent coding framework that encodes an intermediate Gaussian-generation feature before it is mapped to 3D Gaussian primitives. The framework combines entropy-coded latent feature compression, depth-guided multi-view feature refinement, and self-contained latent-to-Gaussian decoding.

  \item We evaluate CodecSplat on DL3DV and RealEstate10K. The proposed method achieves KB-level scene representations and reduces scene size by roughly one order of magnitude compared with compressing feed-forward generated Gaussian primitives.
\end{itemize}

\section{Related Work}
\label{sec:related_work}

\paragraph{Feed-forward 3D Gaussian splatting.}
3DGS represents a scene with explicit Gaussian primitives and enables real-time radiance-field rendering through differentiable rasterization~\citep{kerbl2023gaussians}.
To avoid per-scene optimization, recent feed-forward methods predict Gaussian representations directly from sparse images or posed context views, improving scalability for fast reconstruction and novel-view synthesis.
Representative methods predict Gaussians from single images, image pairs, or sparse multi-view inputs, and improve reconstruction with multi-view stereo cues, depth estimation, recurrent refinement, or large feed-forward Gaussian reconstruction models~\citep{szymanowicz2024splatterimage,charatan2024pixelsplat,chen2024mvsplat,xu2025depthsplat,xu2026resplat,tang2024lgm,xu2024grm,ren2026TokenGS}.
These works mainly focus on reconstruction quality, generalization, and inference efficiency.
CodecSplat addresses a complementary problem: how to produce a compact and decodable scene representation for feed-forward 3DGS under a shared decoder, so that generated scenes can be efficiently stored or transmitted.

\paragraph{Compression of Gaussian representations.}
The explicit nature of 3DGS brings high rendering efficiency but also substantial storage overhead, since a scene may contain a large number of Gaussians with heterogeneous attributes.
Existing 3DGS compression methods mainly reduce this cost through pruning, masking, vector quantization, entropy coding, rate--distortion optimization, or compact Gaussian parameterizations~\citep{lee2024compactgaussian,girish2024eagles,fang2024minisplatting,ali2024trimming,fan2024lightgaussian,navaneet2024compact3dgs,wang2024rdogaussian}.
Other methods further exploit structured anchors, spatial or hash-grid contexts, self-organized attribute layouts, post-training attribute transformation, standard image/video codecs, or compressed Gaussian primitives for more efficient storage~\citep{lu2024scaffoldgs,chen2024hac,wang2024contextgs,morgenstern2024sogs,xie2024mesongs,lee2025codecgs,liu2024compgs}.
More closely related to our setting, FCGS studies fast compression of feed-forward generated 3DGS outputs with learned entropy models over Gaussian attributes~\citep{chen2025fcgs}.
Nevertheless, these methods still compress explicit Gaussian primitives, anchors, or optimized Gaussian parameterizations after they have been formed.
In contrast, CodecSplat moves the coding point inside the feed-forward image-to-Gaussian pipeline and entropy-codes an intermediate Gaussian-generation feature before 2D-to-3D Gaussian construction, allowing compression to exploit a regular, view-aligned, and geometry-aware representation.

\paragraph{Learned latent compression.}
Learned compression maps an input signal to quantized latent representations and encodes the latent symbols with learned entropy models.
End-to-end optimized image codecs established nonlinear transform coding with learned entropy models, where hyperpriors and autoregressive contexts are used to estimate the probability distribution of quantized latents~\citep{balle2017end,balle2018scale,minnen2018joint}.
To support multiple rate points with a single network, variable-rate codecs condition transforms, entropy models, or gain units on rate parameters or quality embeddings~\citep{choi2019variable,cui2021gainedvae,cai2022hifivric}.
CodecSplat follows the principle of entropy-coded latent representation learning, but applies it to a different coding target from conventional image codecs.
Rather than reconstructing RGB images, the decoded latent is optimized to support depth recovery, Gaussian attribute prediction, and 2D-to-3D mapping inside a feed-forward 3DGS pipeline.

\section{Method}
\label{sec:method}

\subsection{Overview}
\label{subsec:overview}

Existing 3DGS compressors typically start from explicit Gaussian primitives, which lie in an irregular 3D space and contain heterogeneous attributes.
In FF3DGS, Gaussian primitives are the outputs of a learned image-to-Gaussian pipeline. Before Gaussian attributes are predicted, the network has organized scene-specific geometry and appearance cues into dense, view-aligned features on regular 2D grids. Since compression seeks compact and structured representations that expose redundancy, we encode this intermediate 2D feature representation rather than the final irregular Gaussian primitives. This gives the codec a compact and regular target that is directly tied to the feed-forward Gaussian generation process.

Given $N$ posed context views
$\mathcal{C}=\{(\mathbf{I}_i,K_i,T_i,n_i,f_i)\}_{i=1}^{N}$,
where $\mathbf{I}_i\in\mathbb{R}^{H\times W\times 3}$ is a context image,
$K_i$ and $T_i$ denote camera intrinsics and extrinsics, and
$n_i,f_i$ are near and far bounds, CodecSplat produces a self-contained scene bitstream $\mathcal{B}$.
The decoder reconstructs from $\mathcal{B}$ a Gaussian representation $\hat{\mathcal{G}}$ that can be rendered from target cameras.

\begin{figure}[t]
  \centering
  \includegraphics[width=1.0\linewidth,trim=10 160 290 10,clip]{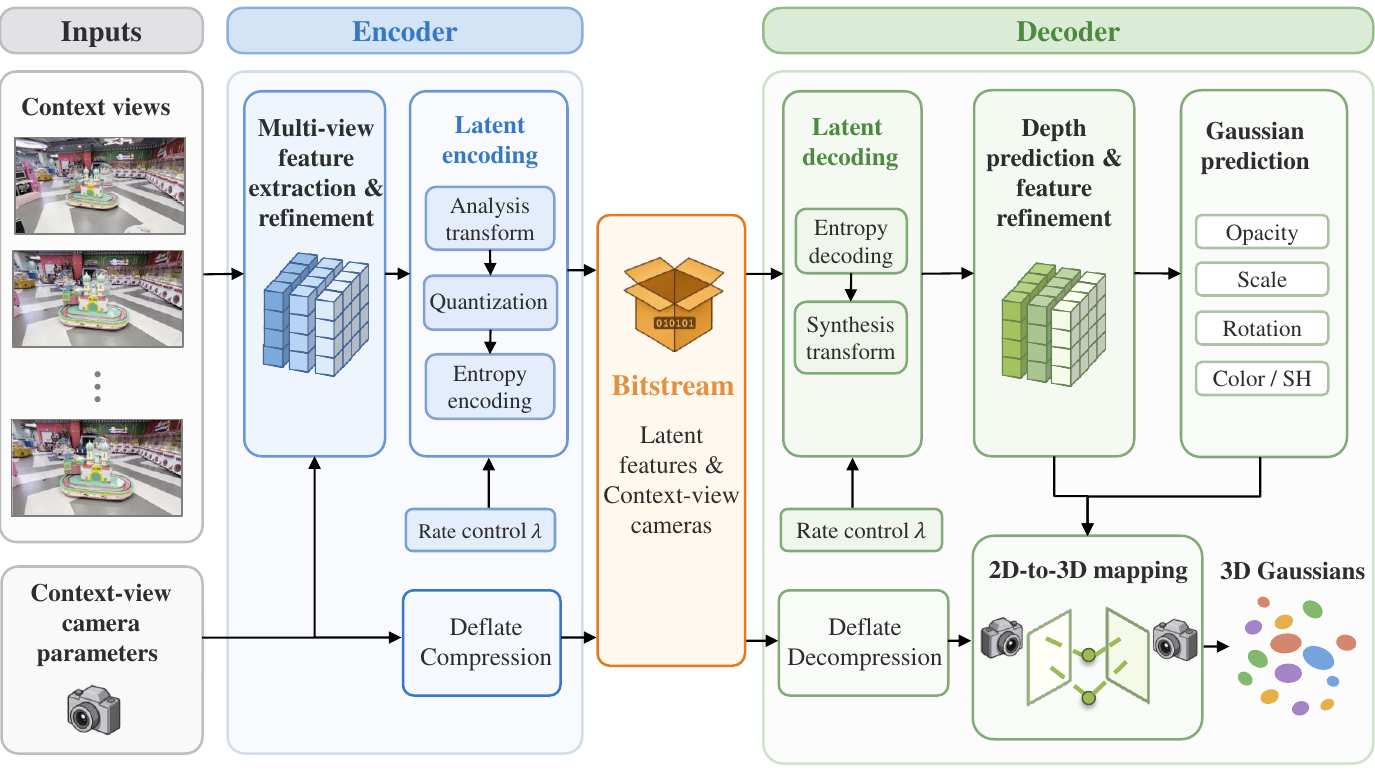}
  \vspace{-1em}
  \caption{Main architecture of CodecSplat.}
  \vspace{-0.5em}
  \label{fig:overview}
\end{figure}

The overall pipeline is illustrated in Fig.~\ref{fig:overview}. The encoder first performs multi-view matching and feature extraction, then compresses the resulting feed-forward feature by latent encoding, including analysis transform, quantization, and entropy coding. The encoded latent features and compressed context-view cameras are packed into a scene bitstream. On the decoder side, CodecSplat entropy-decodes the latent representation, reconstructs the feature with a synthesis transform, and recovers the camera metadata. The decoded feature is then used for depth prediction, feature refinement, Gaussian parameter prediction, and 2D-to-3D mapping.

\subsection{Feed-forward Gaussian-generation representation}
\label{subsec:ff_feature}

CodecSplat is built on a DepthSplat-style FF3DGS backbone~\citep{xu2025depthsplat}.
Given posed context views $\mathcal{C}$, the backbone extracts multi-view features and predicts context-view depth:
\begin{equation}
  (F_{\mathrm{mv}}, D_{\mathrm{enc}}, P_{\mathrm{depth}})
  =
  \operatorname{MVEnc}(\mathcal{C}),
  \label{eq:depthsplat_backbone}
\end{equation}
where $F_{\mathrm{mv}}$ denotes image and cross-view matching features,
$D_{\mathrm{enc}}\in\mathbb{R}^{N\times H\times W}$ is the encoder-side depth prediction, and
$P_{\mathrm{depth}}$ contains depth matching-probability features.
These cues are converted into a dense Gaussian-generation feature:
\begin{equation}
  F_{\mathrm{g}}
  =
  \operatorname{FeatHead}
  \left(\mathbf{I}_{1:N}, F_{\mathrm{mv}}, P_{\mathrm{depth}}\right),
  \qquad
  F_{\mathrm{g}}\in\mathbb{R}^{N\times C_F\times H\times W}.
  \label{eq:gaussian_generation_feature}
\end{equation}

In the codec-free backbone, $F_{\mathrm{g}}$ is consumed by Gaussian prediction heads together with depth to produce explicit Gaussian primitives. CodecSplat instead places the entropy-coded bottleneck on this intermediate feature. The feature is already specialized for Gaussian generation, but it remains a regular view-aligned tensor before 2D-to-3D mapping, making it a more compact and structured coding target than the final unordered primitive set.

\subsection{Depth-guided multi-view feature refinement}
\label{subsec:mv_refinement}

Before latent feature coding, CodecSplat refines the Gaussian-generation feature with depth-guided multi-view interaction.
The refinement has two purposes.
First, depth provides an explicit geometric anchor for Gaussian attribute prediction, since each pixel-aligned prediction will later be mapped to a 3D location.
Second, overlapping context views contain redundant information.
We therefore exchange geometrically aligned features across views before coding, which provides a lightweight form of cross-view redundancy reduction without a multi-view conditional entropy model.

Given a feature tensor $F$ and a depth map $D$, we denote by $A(F,D)$ the geometry-aligned multi-view feature.
For each reference view $r$, it is obtained by warping neighboring source-view features to the reference view, fusing each view pair, and aggregating valid pairwise features:
\begin{equation}
  A_r(F,D)
  =
  \operatorname{Avg}_{s\neq r}
  \left[
    \Phi\left(
    F_r,
    \operatorname{Warp}_{s\rightarrow r}
    \left(F_s; D_r,K_r,T_r,K_s,T_s\right),
    E(d_{r\rightarrow s}^{-1})
    \right)
  \right],
  \label{eq:aligned_feature}
\end{equation}
where $\operatorname{Warp}_{s\rightarrow r}$ back-projects reference-view pixels using the reference depth $D_r$, projects them to the source view $s$, and bilinearly samples $F_s$.
Here $d_{r\rightarrow s}$ is the depth of the back-projected reference point in the source camera, $E$ encodes its inverse depth, and $\Phi$ performs learned pairwise feature fusion. The operator $\operatorname{Avg}$ averages valid pairwise features.

The pre-codec refiner then produces the feature to be compressed by a residual update:
\begin{equation}
  F_{\mathrm c}
  =
  F_{\mathrm g}
  +
  R_{\mathrm{pre}}
  \left(
    F_{\mathrm g},
    A(F_{\mathrm g},D_{\mathrm{enc}})
  \right).
  \label{eq:pre_refine}
\end{equation}
And the same refinement process is used after latent decoding.
Once the reconstructed feature $\hat F_{\mathrm c}$ and codec-side depth $\hat D$ are available, the post-codec refiner produces
\begin{equation}
  F_{\mathrm{dec}}
  =
  \hat F_{\mathrm c}
  +
  R_{\mathrm{post}}
  \left(
    \hat F_{\mathrm c},
    A(\hat F_{\mathrm c},\hat D)
  \right).
  \label{eq:post_refine}
\end{equation}
The pre-codec refiner prepares a more geometry-aware and cross-view informed feature for compression, while the post-codec refiner compensates for compression artifacts before Gaussian prediction.

\subsection{Latent feature coding}
\label{subsec:latent_codec}

The refined feature $F_{\mathrm{c}}$ is compressed by a learned feature codec. A rate parameter $\lambda$ selects the rate point of a single model, and $e_\lambda$ denotes its embedding. The codec consists of an analysis transform, quantization, an entropy model, and a synthesis transform:
\begin{equation}
  Z = \operatorname{Analysis}(F_{\mathrm{c}}; e_\lambda),
  \qquad
  \hat Z = Q(Z),
  \qquad
  \hat F_{\mathrm{c}} = \operatorname{Synthesis}(\hat Z; e_\lambda).
  \label{eq:feature_codec}
\end{equation}
Here $Z=\{Z_\ell\}$ denotes hierarchical latent tensors, $\hat Z$ denotes their quantized symbols, and $\hat F_{\mathrm{c}}$ is the reconstructed feature used by the decoder. We implement this codec as a hierarchical variational autoencoder with learned entropy models~\citep{duan2023lossyvae}.

During training, the rate is estimated by the entropy model and normalized by the feature-map area:
\begin{equation}
  \mathcal{R}
  =
  \frac{1}{HW}
  \sum_{i=1}^{N}
  \sum_{\ell}
  \sum_{u}
  -\log_2
  p_\ell
  \left(
    \hat Z_{i,\ell,u}
    \mid
    C_{i,\ell,u}, e_\lambda
  \right),
  \label{eq:rate}
\end{equation}
where $i$ indexes context views, $\ell$ indexes latent blocks, $u$ indexes spatial/channel positions, and $C_{i,\ell,u}$ is the entropy-model context for the corresponding symbol.

During evaluation, the reported scene size is measured from the packed bitstream, which contains compressed camera metadata and per-view feature streams:
\begin{equation}
  \mathcal{B}
  =
  \operatorname{Pack}
  \left(
    \mathcal{B}_{\mathrm{cam}},
    \{\mathcal{B}^{\mathrm{feat}}_i\}_{i=1}^{N}
  \right),
  \label{eq:bitstream_pack}
\end{equation}
where the camera stream stores the image shape, number of context views, intrinsics, extrinsics, and near/far bounds, and is compressed with Deflate. Each feature stream stores the entropy-coded quantized latents for one context view.

\subsection{Depth and Gaussian decoding}
\label{subsec:latent_to_gaussian}

CodecSplat decodes explicit Gaussian primitives from the reconstructed feature. Since Gaussian centers require 3D geometry, the decoder predicts depth from $\hat F_{\mathrm{c}}$ instead of relying on encoder-side depth or original context images. The depth head predicts a normalized inverse-depth map:
\begin{equation}
  S_i
  =
  \sigma
  \left(
    \operatorname{DepthHead}(\hat F_{\mathrm{c},i})
  \right),
  \qquad
  S_i\in[0,1]^{H\times W}.
  \label{eq:norm_inv_depth}
\end{equation}
The bitstream contains the near and far depth bounds $n_i$ and $f_i$ for each context view.
For each pixel $p$, we convert the normalized inverse-depth value to metric depth by
\begin{equation}
  \hat D_i(p)
  =
  \left[
    f_i^{-1}
    +
    S_i(p)
    \left(n_i^{-1}-f_i^{-1}\right)
  \right]^{-1}.
  \label{eq:bounded_depth}
\end{equation}
Thus $S_i(p)=0$ corresponds to the far bound $f_i$, and $S_i(p)=1$ corresponds to the near bound $n_i$.
This inverse-depth parameterization constrains the decoded depth to the valid scene range.

The post-codec refiner in Eq.~\eqref{eq:post_refine} produces $F_{\mathrm{dec}}$. A Gaussian head predicts pixel-aligned attributes:
\begin{equation}
  \hat\Theta_i
  =
  \operatorname{GSHead}
  \left(F_{\mathrm{dec},i},\hat D_i\right).
  \label{eq:gaussian_params}
\end{equation}
The decoded depth and camera metadata then map the pixel-aligned predictions to 3D Gaussian primitives:
\begin{equation}
  \hat{\mathcal{G}}
  =
  \operatorname{Map}_{2\mathrm{D}\rightarrow3\mathrm{D}}
  \left(
    \{\hat D_i,\hat\Theta_i,K_i,T_i\}_{i=1}^{N}
  \right).
  \label{eq:map_to_3d}
\end{equation}
Here $\operatorname{Map}_{2\mathrm{D}\rightarrow3\mathrm{D}}$ uses the decoded depth and camera parameters to determine Gaussian centers, and instantiates each primitive with the corresponding pixel-aligned attributes.

\subsection{Two-stage optimization}
\label{subsec:two_stage_training}

Directly training CodecSplat with a quantized bottleneck is unstable, as the model must adapt the Gaussian generation pipeline while simultaneously learning a compact latent representation.
We therefore separate the optimization into two stages: a codec-free stage that stabilizes Gaussian generation, followed by a codec-aware stage that optimizes rate--distortion performance.

\paragraph{Stage 1: codec-free Gaussian generation.}
We first train the modified feed-forward Gaussian generation pipeline without the feature codec.
Let $\mathcal{G}_{\mathrm{cf}}$ denote the codec-free Gaussian set predicted from the refined feature $F_{\mathrm{c}}$ and the encoder-side depth $D_{\mathrm{enc}}$.
The target view is rendered as
\begin{equation}
  \hat{\mathbf{Y}}^{\mathrm{cf}}_t
  =
  \operatorname{Render}
  \left(
    \mathcal{G}_{\mathrm{cf}},K_t,T_t
  \right).
  \label{eq:stage1_render}
\end{equation}
The Stage-1 objective is
\begin{equation}
  \mathcal{L}_{\mathrm{stage1}}
  =
  \mathcal{D}_{\mathrm{render}}
  \left(
    \hat{\mathbf{Y}}^{\mathrm{cf}}_t,\mathbf{Y}_t
  \right),
  \label{eq:stage1_loss}
\end{equation}
where $\mathbf{Y}_t$ is the target-view image and $\mathcal{D}_{\mathrm{render}}$ denotes the rendering reconstruction loss.

\paragraph{Stage 2: codec-aware latent coding.}
We initialize from the Stage-1 checkpoint, enable the feature codec and the post-codec refiner, and optimize the codec-related modules while keeping the remaining backbone modules fixed.
After latent coding and Gaussian decoding, the decoded Gaussian set $\hat{\mathcal{G}}$ is rendered to the target view:
\begin{equation}
  \hat{\mathbf{Y}}_t
  =
  \operatorname{Render}
  \left(
    \hat{\mathcal{G}},K_t,T_t
  \right).
  \label{eq:stage2_render}
\end{equation}
The codec-stage objective is
\begin{equation}
  \mathcal{L}_{\mathrm{stage2}}(\lambda)
  =
  \frac{\lambda}{\lambda_{\max}}
  \left[
    \mathcal{D}_{\mathrm{render}}
    \left(
      \hat{\mathbf{Y}}_t,\mathbf{Y}_t
    \right)
    +
    \gamma\,
    \mathcal{D}_{\mathrm{codec\text{-}depth}}
  \right]
  +
  \beta\mathcal{R},
  \label{eq:stage2_loss}
\end{equation}
where $\mathcal{R}$ is the estimated rate, $\gamma$ weights the codec-side depth supervision, and $\beta$ controls the rate penalty.
The codec-side depth loss is defined on normalized inverse depth:
\begin{equation}
  \mathcal{D}_{\mathrm{codec\text{-}depth}}
  =
  \left\|
    S-S_{\mathrm{enc}}
  \right\|_1,
  \label{eq:codec_depth_loss}
\end{equation}
where $S$ is the decoder-side normalized inverse-depth map from Eq.~\eqref{eq:norm_inv_depth}.
The supervision target $S_{\mathrm{enc}}$ is obtained by converting the encoder-side depth $D_{\mathrm{enc}}$ to the same normalized inverse-depth domain:
\begin{equation}
  S_{\mathrm{enc},i}(p)
  =
  \frac{
    D_{\mathrm{enc},i}^{-1}(p)-f_i^{-1}
  }{
    n_i^{-1}-f_i^{-1}
  },
  \label{eq:encoder_depth_to_normalized}
\end{equation}
where $n_i$ and $f_i$ are the near and far bounds of context view $i$.
During training, we sample $\lambda\in[16,1024]$.
At test time, we fix $\lambda$ to obtain a specific rate point.

\section{Experiments}
\label{sec:experiments}

\subsection{Experimental setup}
\label{subsec:exp_setup}

\paragraph{Datasets.}
We evaluate CodecSplat on two standard feed-forward 3D Gaussian splatting benchmarks.
For DL3DV~\citep{liang2024dl3dv}, we follow the ReSplat~\citep{xu2026resplat} 8-input-view benchmark protocol at $256\times448$ resolution.
For RealEstate10K (RE10K)~\citep{zhou2018re10k}, we use the common 2-input-view protocol at $256\times256$ resolution~\citep{xu2025depthsplat,xu2026resplat}.
We report PSNR, SSIM, and LPIPS on target-view rendering.

\paragraph{Training setup.}
CodecSplat is trained in two stages.
The first stage trains the feed-forward Gaussian prediction model without the codec, initialized from the official DepthSplat checkpoint.
The second stage initializes from the first-stage checkpoint and trains the learned feature codec for rate--distortion performance.
The first-stage model is trained for 100K steps, and the second-stage model is trained for 200K steps.
We train all models using bfloat16 mixed precision and the AdamW optimizer, with an initial learning rate of $10^{-4}$ and a weight decay of $0.01$. 
We used four NVIDIA RTX 5880 48GB GPUs.
The learned codec is trained with $\lambda\in[16,1024]$ for rate--distortion control.
In the second-stage objective, we set the codec-side depth loss weight to $\gamma=0.1$, and use the rate weight $\beta=10^{-4}$ for DL3DV and $\beta=4\times10^{-4}$ for RE10K.

\paragraph{Baselines.}
We compare against released DepthSplat and ReSplat models as feed-forward reconstruction references.
DepthSplat predicts pixel-aligned Gaussians from sparse views with explicit depth estimation, while ReSplat improves feed-forward Gaussian prediction with recurrent refinement.
To evaluate compressors that operate on generated Gaussian primitives, we first export complete Gaussian PLY files from DepthSplat and ReSplat, then apply existing 3DGS compressors to the exported Gaussian primitives.
We consider FCGS, gsplat PNG Compression, and SOGS~\citep{chen2025fcgs,ye2025gsplat,morgenstern2024sogs}.
FCGS performs learned compression of feed-forward generated Gaussian attributes.
We use released FCGS models at $\lambda_{\mathrm{FCGS}}\in\{0.0001,0.0004,0.0016\}$.
gsplat PNG Compression quantizes and stores Gaussian attributes with standard image-style coding.
We use $K_{\mathrm{SH}}\in\{128,8192\}$ clusters for high-order spherical harmonic coefficients.
SOGS reorganizes Gaussian parameters into locally smooth 2D grids before compression.
We use its default compression configuration as a reference. For lower-rate tests on ReSplat outputs, we modify PNG/SOGS quantization from the default 16/8-bit means/attribute precision to 4/2, 8/4, and 12/6 bits.
In the tables, $A\rightarrow B$ means that backbone $A$ first generates Gaussian primitives and codec $B$ then compresses the generated output.

\paragraph{Bitstream accounting and timing.}
For CodecSplat, the reported scene size includes camera metadata and per-view feature streams.
Context cameras are included for CodecSplat because they are required for decoder-side mapping from 2D predictions to 3D Gaussian primitives.
For timing, we report \emph{GS generation}, \emph{compression}, and \emph{decompression}.
Compression and decompression correspond to codec encode/decode for both CodecSplat and baselines that compress generated primitives.
GS generation includes the feed-forward generation needed before compression.
For baselines that compress generated primitives, GS generation includes generating and preparing the Gaussian representation.
For CodecSplat, GS generation corresponds to producing the intermediate representation before latent feature coding.

\begin{figure}[t]
  \centering
  \includegraphics[width=1.0\linewidth,trim=0 569 19 0,clip]{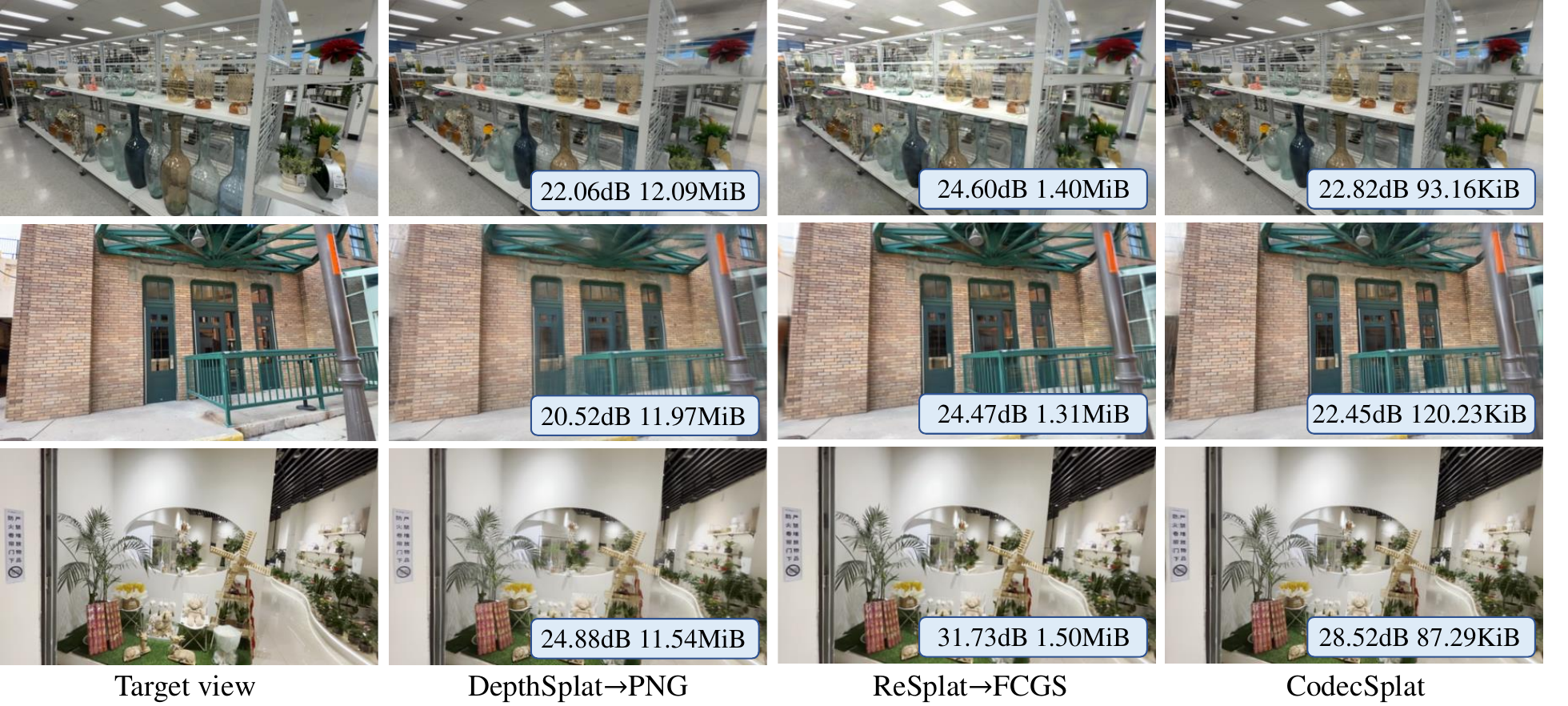}
  \vspace{-1.5em}
  \caption{Qualitative comparison on target-view rendering. We compare CodecSplat with Gaussian-output compression baselines that first generate explicit Gaussian primitives and then compress them. Compared with MiB-level compressed Gaussian outputs, CodecSplat produces KiB-level scene bitstreams while preserving the dominant scene structure and overall appearance.
}
\vspace{-0.8em}
  \label{fig:vis}
\end{figure}

\subsection{Main compression results}
\label{subsec:main_results}

{
\begin{table}[t]
\centering
\caption{Rate-distortion comparison with methods that compress generated feed-forward Gaussian primitives. For PNG/SOGS, the codec setting lists the means/attribute quantization bits. PNG uses $K_{\mathrm{SH}}=8192$ unless otherwise specified.}
\label{tab:main_comparison}
\renewcommand{\arraystretch}{0.85}
\setlength{\tabcolsep}{6.1pt}
\footnotesize
\begin{tabular}{cllrrrr}
\toprule
Dataset & Method & Codec Setting & KiB/scene $\downarrow$ & PSNR $\uparrow$ & SSIM $\uparrow$ & LPIPS $\downarrow$ \\
\midrule
\multirow{22}{*}{DL3DV}
    & DepthSplat$\rightarrow$PNG & 16/8 bits & 11942.05 & 24.94 & 0.848 & 0.128 \\
    & DepthSplat$\rightarrow$SOGS & 16/8 bits & 11759.53 & 24.94 & 0.848 & 0.128 \\
    & DepthSplat$\rightarrow$FCGS & $\lambda_{\mathrm{FCGS}}=0.0016$ & 12367.10 & 24.93 & 0.847 & 0.129 \\
    & ReSplat$\rightarrow$PNG & 4/2 bits & 282.74 & 8.67 & 0.298 & 0.818 \\
    & ReSplat$\rightarrow$PNG & 8/4 bits & 447.16 & 19.79 & 0.639 & 0.404 \\
    & ReSplat$\rightarrow$PNG & 12/6 bits & 659.55 & 22.62 & 0.833 & 0.188 \\
    & ReSplat$\rightarrow$PNG & 16/8 bits & 856.74 & 22.75 & 0.843 & 0.172 \\
    & ReSplat$\rightarrow$PNG & 16/8 bits, $K_{\mathrm{SH}}=128$ & 741.09 & 23.36 & 0.841 & 0.171 \\
    & ReSplat$\rightarrow$SOGS & 4/2 bits & 466.49 & 11.85 & 0.348 & 0.764 \\
    & ReSplat$\rightarrow$SOGS & 8/4 bits & 604.84 & 23.46 & 0.727 & 0.267 \\
    & ReSplat$\rightarrow$SOGS & 12/6 bits & 759.27 & 28.27 & 0.892 & 0.123 \\
    & ReSplat$\rightarrow$SOGS & 16/8 bits & 879.45 & 28.36 & 0.894 & 0.120 \\
    & ReSplat$\rightarrow$FCGS & $\lambda_{\mathrm{FCGS}}=0.0016$ & 1425.22 & 28.72 & 0.897 & 0.113 \\
    & ReSplat$\rightarrow$FCGS & $\lambda_{\mathrm{FCGS}}=0.0004$ & 1641.63 & 28.88 & 0.900 & 0.109 \\
    & ReSplat$\rightarrow$FCGS & $\lambda_{\mathrm{FCGS}}=0.0001$ & 1926.43 & 28.93 & 0.901 & 0.108 \\
    & CodecSplat & $\lambda=16$ & 20.00 & 23.56 & 0.768 & 0.267 \\
    & CodecSplat & $\lambda=32$ & 28.70 & 24.48 & 0.809 & 0.209 \\
    & CodecSplat & $\lambda=64$ & 41.52 & 25.24 & 0.839 & 0.167 \\
    & CodecSplat & $\lambda=128$ & 57.79 & 25.79 & 0.858 & 0.141 \\
    & CodecSplat & $\lambda=256$ & 76.56 & 26.14 & 0.869 & 0.127 \\
    & CodecSplat & $\lambda=512$ & 94.38 & 26.30 & 0.874 & 0.120 \\
    & CodecSplat & $\lambda=1024$ & 107.77 & 26.36 & 0.876 & 0.117 \\
\midrule
\multirow{22}{*}{RE10K}
    & DepthSplat$\rightarrow$PNG & 16/8 bits & 1678.17 & 27.28 & 0.885 & 0.117 \\
    & DepthSplat$\rightarrow$SOGS & 16/8 bits & 1670.96 & 27.28 & 0.885 & 0.116 \\
    & DepthSplat$\rightarrow$FCGS & $\lambda_{\mathrm{FCGS}}=0.0016$ & 1869.35 & 26.01 & 0.872 & 0.146 \\
    & ReSplat$\rightarrow$PNG & 4/2 bits & 518.90 & 10.93 & 0.307 & 0.747 \\
    & ReSplat$\rightarrow$PNG & 8/4 bits & 890.08 & 22.38 & 0.754 & 0.318 \\
    & ReSplat$\rightarrow$PNG & 12/6 bits & 1372.35 & 25.66 & 0.880 & 0.133 \\
    & ReSplat$\rightarrow$PNG & 16/8 bits & 1841.51 & 25.90 & 0.887 & 0.122 \\
    & ReSplat$\rightarrow$PNG & 16/8 bits, $K_{\mathrm{SH}}=128$ & 1674.83 & 26.39 & 0.891 & 0.119 \\
    & ReSplat$\rightarrow$SOGS & 4/2 bits & 849.09 & 12.73 & 0.427 & 0.716 \\
    & ReSplat$\rightarrow$SOGS & 8/4 bits & 1119.22 & 26.25 & 0.836 & 0.202 \\
    & ReSplat$\rightarrow$SOGS & 12/6 bits & 1473.14 & 29.33 & 0.907 & 0.104 \\
    & ReSplat$\rightarrow$SOGS & 16/8 bits & 1841.91 & 29.36 & 0.908 & 0.102 \\
    & ReSplat$\rightarrow$FCGS & $\lambda_{\mathrm{FCGS}}=0.0016$ & 2526.32 & 29.16 & 0.905 & 0.109 \\
    & ReSplat$\rightarrow$FCGS & $\lambda_{\mathrm{FCGS}}=0.0004$ & 2959.61 & 29.56 & 0.909 & 0.103 \\
    & ReSplat$\rightarrow$FCGS & $\lambda_{\mathrm{FCGS}}=0.0001$ & 3555.14 & 29.66 & 0.910 & 0.101 \\
    & CodecSplat & $\lambda=16$ & 3.37 & 24.76 & 0.809 & 0.231 \\
    & CodecSplat & $\lambda=32$ & 4.28 & 25.55 & 0.839 & 0.188 \\
    & CodecSplat & $\lambda=64$ & 5.58 & 26.18 & 0.859 & 0.157 \\
    & CodecSplat & $\lambda=128$ & 7.13 & 26.59 & 0.871 & 0.140 \\
    & CodecSplat & $\lambda=256$ & 8.89 & 26.84 & 0.877 & 0.130 \\
    & CodecSplat & $\lambda=512$ & 10.78 & 26.99 & 0.880 & 0.125 \\
    & CodecSplat & $\lambda=1024$ & 12.51 & 27.05 & 0.882 & 0.123 \\
\bottomrule
\end{tabular}
\vspace{-1.5em}
\end{table}
}

{
\begin{table}[t]
\centering
\caption{Timing breakdown in seconds. GS generation includes the feed-forward GS generation stage, and compression/decompression corresponds to codec encode/decode.}
\label{tab:runtime}
\renewcommand{\arraystretch}{0.9}
\footnotesize
\begin{tabular}{cllrrr}
\toprule
Dataset & Method & Codec Setting & GS Gen. T. $\downarrow$ & Compress T. $\downarrow$ & Decompress T. $\downarrow$ \\
\midrule
\multirow{7}{*}{DL3DV}
    & DepthSplat$\rightarrow$PNG & $K_{\mathrm{SH}}=8192$ & 0.155 & 18.277 & 0.487 \\
    & DepthSplat$\rightarrow$SOGS & Default & 0.155 & 69.768 & 0.452 \\
    & DepthSplat$\rightarrow$FCGS & $\lambda_{\mathrm{FCGS}}=0.0016$ & 0.155 & 4.470 & 3.790 \\
    & ReSplat$\rightarrow$PNG & $K_{\mathrm{SH}}=8192$ & 0.282 & 2.919 & 0.050 \\
    & ReSplat$\rightarrow$SOGS & Default & 0.282 & 16.114 & 0.024 \\
    & ReSplat$\rightarrow$FCGS & $\lambda_{\mathrm{FCGS}}=0.0016$ & 0.282 & 0.762 & 0.969 \\
    & CodecSplat & $\lambda=16,\ldots,1024$ & 0.124 & 0.267 & 0.273 \\
\midrule
\multirow{7}{*}{RE10K}
    & DepthSplat$\rightarrow$PNG & $K_{\mathrm{SH}}=8192$ & 0.046 & 2.778 & 0.075 \\
    & DepthSplat$\rightarrow$SOGS & Default & 0.046 & 53.856 & 0.049 \\
    & DepthSplat$\rightarrow$FCGS & $\lambda_{\mathrm{FCGS}}=0.0016$ & 0.046 & 0.851 & 0.928 \\
    & ReSplat$\rightarrow$PNG & $K_{\mathrm{SH}}=8192$ & 0.284 & 2.901 & 0.081 \\
    & ReSplat$\rightarrow$SOGS & Default & 0.284 & 60.284 & 0.062 \\
    & ReSplat$\rightarrow$FCGS & $\lambda_{\mathrm{FCGS}}=0.0016$ & 0.284 & 1.062 & 1.271 \\
    & CodecSplat & $\lambda=16,\ldots,1024$ & 0.047 & 0.062 & 0.059 \\
\bottomrule
\end{tabular}
\end{table}
}

Table~\ref{tab:main_comparison} compares CodecSplat with methods that compress generated feed-forward Gaussian primitives.
The seven CodecSplat rows on each dataset correspond to fixed test-time values of $\lambda$.
Increasing $\lambda$ consistently increases the bitstream size and improves reconstruction quality, showing controllable rate--distortion behavior with a single trained codec.
On DL3DV, CodecSplat ranges from 20.00 to 107.77 KiB per scene and reaches 23.56--26.36 dB PSNR.
On RE10K, CodecSplat uses only 3.37--12.51 KiB per scene while reaching 24.76--27.05 dB PSNR.

The results show a clear difference between compressing generated primitives and coding the internal feed-forward representation.
DepthSplat-based baselines require large bitstreams on both datasets: 11.48--12.08 MiB on DL3DV and 1.63--1.83 MiB on RE10K.
This is because DepthSplat predicts pixel-aligned Gaussians at the input image resolution, producing a dense set of primitives that remains expensive to store even after applying 3DGS compressors.
In contrast, ReSplat generates a more compact Gaussian output by predicting Gaussians on downsampled feature maps and refining them through 3D context extraction and recurrent refinement.
This makes ReSplat more compression-friendly and also improves rendering quality, as shown by ReSplat$\rightarrow$SOGS and ReSplat$\rightarrow$FCGS.
For example, on DL3DV, ReSplat$\rightarrow$FCGS reaches 28.72 dB with 1.39 MiB, and ReSplat$\rightarrow$SOGS reaches 28.36 dB with 0.86 MiB.
The cost of this stronger representation is higher feed-forward generation complexity, as reflected in Table~\ref{tab:runtime}.

CodecSplat follows a different design.
Instead of first generating a full Gaussian output and then compressing it, CodecSplat encodes the intermediate Gaussian-generation feature before it is mapped to 3D Gaussian primitives.
This leads to substantially smaller scene representations.
ReSplat-based SOGS and FCGS outputs retain higher PSNR at high-rate settings, whereas aggressive PNG/SOGS quantization reduces the rate at the cost of substantial quality degradation. CodecSplat provides a much lower bitrate range while retaining competitive reconstruction quality.

{
\begin{table}[t]
\centering
\caption{Codec-free ablation on DL3DV.}
\label{tab:codecfree_ablation}
\footnotesize
\setlength{\tabcolsep}{13.6pt}
\renewcommand{\arraystretch}{0.9}
\begin{tabular}{lrrrr}
\toprule
Method & PSNR $\uparrow$ & SSIM $\uparrow$ & LPIPS $\downarrow$ & GS Gen. T. $\downarrow$ \\
\midrule
DepthSplat & 24.94 & 0.848 & 0.128 & 0.155 \\
ReSplat & 28.96 & 0.901 & 0.107 & 0.282 \\
CodecSplat w/o codec w/o refiner & 25.68 & 0.866 & 0.116 & 0.102 \\
CodecSplat w/o codec & 26.22 & 0.880 & 0.108 & 0.146 \\
\bottomrule
\end{tabular}
\end{table}
}

{
\begin{table}[tb]
\centering
\caption{Naive tensor-compression ablation on DL3DV.}
\label{tab:deflate_ablation}
\footnotesize
\setlength{\tabcolsep}{10.3pt}
\renewcommand{\arraystretch}{0.9}
\begin{tabular}{llrrrr}
\toprule
Method & Codec Setting & Bytes/scene $\downarrow$ & PSNR $\uparrow$ & SSIM $\uparrow$ & LPIPS $\downarrow$ \\
\midrule
CodecSplat & $\lambda=1024$ & 0.11 MiB & 26.36 & 0.876 & 0.117 \\
CodecSplat w/ Deflate & float16 features + depth & 81.08 MiB & 26.22 & 0.880 & 0.108 \\
\bottomrule
\end{tabular}
\vspace{-0.5em}
\end{table}
}

\subsection{Coding-time analysis}
\label{subsec:runtime}

Table~\ref{tab:runtime} reports the timing breakdown.
CodecSplat keeps codec operations lightweight in absolute time: compression/decompression takes 0.267/0.273 seconds on DL3DV and 0.062/0.059 seconds on RE10K.
It avoids the multi-second compression cost observed for several baselines that operate on exported Gaussian files, while achieving subsecond decompression on both datasets.
CodecSplat primarily targets compact scene storage, and its codec time remains practical for the tested settings.

\subsection{Ablation and transfer studies}
\label{subsec:ablation}

\paragraph{Codec-free Gaussian generation.}
Table~\ref{tab:codecfree_ablation} evaluates the codec-free variants on DL3DV.
This ablation serves two purposes.
First, it measures the reconstruction upper bound of each FF3DGS backbone before compression is introduced.
The codec-free CodecSplat reaches 26.22 dB PSNR, outperforming DepthSplat but remaining below ReSplat.
This explains why CodecSplat has lower PSNR than ReSplat-based compression baselines in Table~\ref{tab:main_comparison}: our Gaussian generation backbone follows a DepthSplat-style design with lower model complexity, while ReSplat uses recurrent refinement and stronger 3D feature interaction.
Second, the ablation isolates the effect of the depth-guided multi-view refiner.
Adding the refiner improves the codec-free model from 25.68 dB to 26.22 dB, showing that cross-view feature interaction strengthens the Gaussian-generation representation before coding.
Notably, at $\lambda=1024$, CodecSplat achieves 26.36 dB PSNR in Table~\ref{tab:main_comparison}, matching the codec-free upper bound without observable quality degradation.
This indicates that the proposed rate-control mechanism can approach the reconstruction capacity of the underlying feed-forward Gaussian generator at high bitrate, while still supporting much lower rate points.

\paragraph{Naive tensor compression.}
This ablation examines whether a learned codec is necessary for integrated FF3DGS compression.
A simple alternative is to serialize the intermediate feature representation and predicted depth from the codec-free model in float16 and compress them with Deflate.
This baseline does not use the second-stage codec-aware training of CodecSplat, and serves as a comparison for directly storing the intermediate representation.
As shown in Table~\ref{tab:deflate_ablation}, Deflate preserves the codec-free reconstruction quality but requires 81.08 MiB per scene.
In contrast, CodecSplat achieves 26.36 dB PSNR with only 0.11 MiB at $\lambda=1024$.
This result shows that moving the coding point earlier is not sufficient by itself.
A learned codec is necessary to transform the intermediate representation into a compact latent space while preserving the information needed for Gaussian decoding.

{
\begin{table}[t]
\centering
\caption{Refinement ablation with the learned codec enabled on DL3DV.}
\label{tab:refinement_ablation}
\footnotesize
\setlength{\tabcolsep}{10.9pt}
\renewcommand{\arraystretch}{0.9}
\begin{tabular}{crrrrrr}
\toprule
\multirow{2}{*}{$\lambda$} & \multicolumn{2}{c}{Full CodecSplat} & \multicolumn{2}{c}{Without pre-refiner} & \multicolumn{2}{c}{Without post-refiner} \\
    & KiB/scene $\downarrow$ & PSNR $\uparrow$ & KiB/scene $\downarrow$ & PSNR $\uparrow$ & KiB/scene $\downarrow$ & PSNR $\uparrow$ \\
\midrule
16 & 20.00 & 23.56 & 33.39 & 24.44 & 33.02 & 24.37 \\
32 & 28.70 & 24.48 & 46.66 & 25.21 & 44.64 & 25.04 \\
64 & 41.52 & 25.24 & 63.88 & 25.75 & 61.02 & 25.57 \\
128 & 57.79 & 25.79 & 83.44 & 26.05 & 81.05 & 25.94 \\
256 & 76.56 & 26.14 & 102.98 & 26.22 & 102.42 & 26.17 \\
512 & 94.38 & 26.30 & 119.73 & 26.28 & 119.88 & 26.25 \\
1024 & 107.77 & 26.36 & 130.13 & 26.29 & 131.65 & 26.28 \\
\bottomrule
\end{tabular}
\end{table}
}

{
\begin{table}[t]
\centering
\caption{Transfer to ZPressor on DL3DV.}
\label{tab:zpressor_transfer}
\footnotesize
\setlength{\tabcolsep}{11.3pt}
\renewcommand{\arraystretch}{0.9}
\begin{tabular}{crrrrr}
\toprule
\multirow{2}{*}{Context views}
    & ZPressor & \multicolumn{2}{c}{ZPressor+CodecSplat $\lambda=16$} & \multicolumn{2}{c}{ZPressor+CodecSplat $\lambda=1024$} \\
    & PSNR $\uparrow$ & KiB/scene $\downarrow$ & PSNR $\uparrow$ & KiB/scene $\downarrow$ & PSNR $\uparrow$ \\
\midrule
12 & 24.30 & 17.56 & 22.96 & 85.65 & 24.66 \\
16 & 24.25 & 17.72 & 22.93 & 86.02 & 24.60 \\
24 & 24.26 & 17.77 & 22.90 & 86.25 & 24.59 \\
36 & 23.88 & 17.79 & 22.43 & 86.30 & 24.16 \\
\bottomrule
\end{tabular}
\end{table}
}

\paragraph{Pre- and post-codec refinement.}
We train two codec variants, removing either the pre-codec depth-guided refinement module or the post-codec refinement module.
Table~\ref{tab:refinement_ablation} compares both variants with the full model on DL3DV at seven rate points.
We assess the contribution of refinement through rate-distortion performance across these rate points.
The full model reduces BD-rate by 13.67\% relative to the model without the pre-refiner and by 18.08\% relative to the model without the post-refiner.
Both refinement stages therefore help the codec allocate rate more effectively rather than merely increasing reconstruction capacity.

\paragraph{Transfer to ZPressor.}
To test whether our coding strategy transfers beyond the original pipeline, we integrate CodecSplat into another FF3DGS model, ZPressor~\citep{wang2025zpressor}, which aggregates support-view information into selected anchor-view features before Gaussian prediction.
We train one variable-rate model using ZPressor's 12-view, 6-cluster protocol and evaluate the same model with 12, 16, 24, and 36 context views on DL3DV.
Table~\ref{tab:zpressor_transfer} reports the results at $\lambda=16$ and $\lambda=1024$, alongside the original ZPressor PSNR values.
At $\lambda=1024$, ZPressor+CodecSplat exceeds the published PSNR by 0.28--0.36 dB while using 85.65--86.30 KiB per scene.
This shows that the latent coding strategy transfers to a substantially different multi-view feature aggregation pipeline, achieving KiB-scale representations while preserving reconstruction quality.

\section{Conclusion}
\label{sec:conclusion}
We presented CodecSplat, an ultra-compact latent coding framework for feed-forward 3D Gaussian splatting. Instead of compressing generated Gaussian primitives, CodecSplat encodes an intermediate Gaussian-generation feature and decodes it into Gaussian primitives, allowing the codec to operate on a compact and regular 2D representation before it is mapped to irregular 3D primitives.
Experiments on DL3DV and RE10K demonstrate KB-level scene representations with controllable rate--distortion behavior, suggesting that the intermediate representation inside FF3DGS provides an effective coding point for compact storage and transmission.

\paragraph{Limitations and future work.} CodecSplat prioritizes compactness over peak novel-view synthesis quality, and its PSNR is still limited by the underlying FF3DGS backbone and the current latent-to-Gaussian decoder.
Since the codec is integrated into the feed-forward Gaussian generation pipeline, it also requires joint training rather than being directly plugged into arbitrary pretrained FF3DGS models.
Future work will explore stronger backbones and more expressive decoding designs to improve reconstruction quality while preserving compact scene representations.

\section*{Acknowledgements}
This work was partially supported by the National Natural Science Foundation of China (No. 62372491), the Guangdong S\&T Programme (No. 2025B0101130003), the Guangdong Basic and Applied Basic Research Foundation (2023B1515120087), and the Science and Technology Planning Project of Key Laboratory of Advanced IntelliSense Technology, Guangdong Science and Technology Department (2023B1212060024), and the Major Key Project of Pengcheng Laboratory (No. PCL2026A02).

\bibliographystyle{plainnat}
\bibliography{reference}

\input{checklist}
\end{document}

%% file: checklist.tex
\section*{NeurIPS Paper Checklist}

\begin{enumerate}

\item {\bf Claims}
    \item[] Question: Do the main claims made in the abstract and introduction accurately reflect the paper's contributions and scope?
    \item[] Answer: \answerYes{} 
    \item[] Justification: The abstract and introduction state the integrated FF3DGS compression problem, the CodecSplat framework, and the empirical claim of compact scene representations. These claims are supported by the method description and the experiments on DL3DV and RealEstate10K.
    \item[] Guidelines:
    \begin{itemize}
        \item The answer \answerNA{} means that the abstract and introduction do not include the claims made in the paper.
        \item The abstract and/or introduction should clearly state the claims made, including the contributions made in the paper and important assumptions and limitations. A \answerNo{} or \answerNA{} answer to this question will not be perceived well by the reviewers. 
        \item The claims made should match theoretical and experimental results, and reflect how much the results can be expected to generalize to other settings. 
        \item It is fine to include aspirational goals as motivation as long as it is clear that these goals are not attained by the paper. 
    \end{itemize}

\item {\bf Limitations}
    \item[] Question: Does the paper discuss the limitations of the work performed by the authors?
    \item[] Answer: \answerYes{} 
    \item[] Justification: The conclusion discusses the remaining novel-view synthesis quality gap, the dependence on the FF3DGS backbone and latent-to-Gaussian decoder, and the need for joint training with the feed-forward pipeline.
    \item[] Guidelines:
    \begin{itemize}
        \item The answer \answerNA{} means that the paper has no limitation while the answer \answerNo{} means that the paper has limitations, but those are not discussed in the paper. 
        \item The authors are encouraged to create a separate ``Limitations'' section in their paper.
        \item The paper should point out any strong assumptions and how robust the results are to violations of these assumptions (e.g., independence assumptions, noiseless settings, model well-specification, asymptotic approximations only holding locally). The authors should reflect on how these assumptions might be violated in practice and what the implications would be.
        \item The authors should reflect on the scope of the claims made, e.g., if the approach was only tested on a few datasets or with a few runs. In general, empirical results often depend on implicit assumptions, which should be articulated.
        \item The authors should reflect on the factors that influence the performance of the approach. For example, a facial recognition algorithm may perform poorly when image resolution is low or images are taken in low lighting. Or a speech-to-text system might not be used reliably to provide closed captions for online lectures because it fails to handle technical jargon.
        \item The authors should discuss the computational efficiency of the proposed algorithms and how they scale with dataset size.
        \item If applicable, the authors should discuss possible limitations of their approach to address problems of privacy and fairness.
        \item While the authors might fear that complete honesty about limitations might be used by reviewers as grounds for rejection, a worse outcome might be that reviewers discover limitations that aren't acknowledged in the paper. The authors should use their best judgment and recognize that individual actions in favor of transparency play an important role in developing norms that preserve the integrity of the community. Reviewers will be specifically instructed to not penalize honesty concerning limitations.
    \end{itemize}

\item {\bf Theory assumptions and proofs}
    \item[] Question: For each theoretical result, does the paper provide the full set of assumptions and a complete (and correct) proof?
    \item[] Answer: \answerNA{} 
    \item[] Justification: The paper does not present theoretical theorems or proof-based results.
    \item[] Guidelines:
    \begin{itemize}
        \item The answer \answerNA{} means that the paper does not include theoretical results. 
        \item All the theorems, formulas, and proofs in the paper should be numbered and cross-referenced.
        \item All assumptions should be clearly stated or referenced in the statement of any theorems.
        \item The proofs can either appear in the main paper or the supplemental material, but if they appear in the supplemental material, the authors are encouraged to provide a short proof sketch to provide intuition. 
        \item Inversely, any informal proof provided in the core of the paper should be complemented by formal proofs provided in appendix or supplemental material.
        \item Theorems and Lemmas that the proof relies upon should be properly referenced. 
    \end{itemize}

    \item {\bf Experimental result reproducibility}
    \item[] Question: Does the paper fully disclose all the information needed to reproduce the main experimental results of the paper to the extent that it affects the main claims and/or conclusions of the paper (regardless of whether the code and data are provided or not)?
    \item[] Answer: \answerYes{} 
    \item[] Justification: The paper describes the model pipeline, two-stage training, datasets, evaluation protocols, rate-control settings, optimizer, loss weights, baselines, and bitstream accounting needed to reproduce the main results.
    \item[] Guidelines:
    \begin{itemize}
        \item The answer \answerNA{} means that the paper does not include experiments.
        \item If the paper includes experiments, a \answerNo{} answer to this question will not be perceived well by the reviewers: Making the paper reproducible is important, regardless of whether the code and data are provided or not.
        \item If the contribution is a dataset and\slash or model, the authors should describe the steps taken to make their results reproducible or verifiable. 
        \item Depending on the contribution, reproducibility can be accomplished in various ways. For example, if the contribution is a novel architecture, describing the architecture fully might suffice, or if the contribution is a specific model and empirical evaluation, it may be necessary to either make it possible for others to replicate the model with the same dataset, or provide access to the model. In general. releasing code and data is often one good way to accomplish this, but reproducibility can also be provided via detailed instructions for how to replicate the results, access to a hosted model (e.g., in the case of a large language model), releasing of a model checkpoint, or other means that are appropriate to the research performed.
        \item While NeurIPS does not require releasing code, the conference does require all submissions to provide some reasonable avenue for reproducibility, which may depend on the nature of the contribution. For example
        \begin{enumerate}
            \item If the contribution is primarily a new algorithm, the paper should make it clear how to reproduce that algorithm.
            \item If the contribution is primarily a new model architecture, the paper should describe the architecture clearly and fully.
            \item If the contribution is a new model (e.g., a large language model), then there should either be a way to access this model for reproducing the results or a way to reproduce the model (e.g., with an open-source dataset or instructions for how to construct the dataset).
            \item We recognize that reproducibility may be tricky in some cases, in which case authors are welcome to describe the particular way they provide for reproducibility. In the case of closed-source models, it may be that access to the model is limited in some way (e.g., to registered users), but it should be possible for other researchers to have some path to reproducing or verifying the results.
        \end{enumerate}
    \end{itemize}

\item {\bf Open access to data and code}
    \item[] Question: Does the paper provide open access to the data and code, with sufficient instructions to faithfully reproduce the main experimental results, as described in supplemental material?
    \item[] Answer: \answerYes{} 
    \item[] Justification: Code is available at \url{https://github.com/pengpeng-yu/CodecSplat}.
    \item[] Guidelines:
    \begin{itemize}
        \item The answer \answerNA{} means that paper does not include experiments requiring code.
        \item Please see the NeurIPS code and data submission guidelines (\url{https://neurips.cc/public/guides/CodeSubmissionPolicy}) for more details.
        \item While we encourage the release of code and data, we understand that this might not be possible, so \answerNo{} is an acceptable answer. Papers cannot be rejected simply for not including code, unless this is central to the contribution (e.g., for a new open-source benchmark).
        \item The instructions should contain the exact command and environment needed to run to reproduce the results. See the NeurIPS code and data submission guidelines (\url{https://neurips.cc/public/guides/CodeSubmissionPolicy}) for more details.
        \item The authors should provide instructions on data access and preparation, including how to access the raw data, preprocessed data, intermediate data, and generated data, etc.
        \item The authors should provide scripts to reproduce all experimental results for the new proposed method and baselines. If only a subset of experiments are reproducible, they should state which ones are omitted from the script and why.
        \item At submission time, to preserve anonymity, the authors should release anonymized versions (if applicable).
        \item Providing as much information as possible in supplemental material (appended to the paper) is recommended, but including URLs to data and code is permitted.
    \end{itemize}

\item {\bf Experimental setting/details}
    \item[] Question: Does the paper specify all the training and test details (e.g., data splits, hyperparameters, how they were chosen, type of optimizer) necessary to understand the results?
    \item[] Answer: \answerYes{} 
    \item[] Justification: The experimental setup specifies the datasets, input-view protocols, resolutions, two-stage training schedule, optimizer, learning rate, weight decay, rate range, loss weights, baseline settings, and evaluation metrics.
    \item[] Guidelines:
    \begin{itemize}
        \item The answer \answerNA{} means that the paper does not include experiments.
        \item The experimental setting should be presented in the core of the paper to a level of detail that is necessary to appreciate the results and make sense of them.
        \item The full details can be provided either with the code, in appendix, or as supplemental material.
    \end{itemize}

\item {\bf Experiment statistical significance}
    \item[] Question: Does the paper report error bars suitably and correctly defined or other appropriate information about the statistical significance of the experiments?
    \item[] Answer: \answerNo{} 
    \item[] Justification: The paper reports benchmark averages under standard evaluation protocols but does not include repeated runs, confidence intervals, or error bars due to computational cost. The main conclusions are based on consistent rate--distortion trends and large differences in scene size.
    \item[] Guidelines:
    \begin{itemize}
        \item The answer \answerNA{} means that the paper does not include experiments.
        \item The authors should answer \answerYes{} if the results are accompanied by error bars, confidence intervals, or statistical significance tests, at least for the experiments that support the main claims of the paper.
        \item The factors of variability that the error bars are capturing should be clearly stated (for example, train/test split, initialization, random drawing of some parameter, or overall run with given experimental conditions).
        \item The method for calculating the error bars should be explained (closed form formula, call to a library function, bootstrap, etc.)
        \item The assumptions made should be given (e.g., Normally distributed errors).
        \item It should be clear whether the error bar is the standard deviation or the standard error of the mean.
        \item It is OK to report 1-sigma error bars, but one should state it. The authors should preferably report a 2-sigma error bar than state that they have a 96\% CI, if the hypothesis of Normality of errors is not verified.
        \item For asymmetric distributions, the authors should be careful not to show in tables or figures symmetric error bars that would yield results that are out of range (e.g., negative error rates).
        \item If error bars are reported in tables or plots, the authors should explain in the text how they were calculated and reference the corresponding figures or tables in the text.
    \end{itemize}

\item {\bf Experiments compute resources}
    \item[] Question: For each experiment, does the paper provide sufficient information on the computer resources (type of compute workers, memory, time of execution) needed to reproduce the experiments?
    \item[] Answer: \answerYes{} 
    \item[] Justification: We used four NVIDIA RTX 5880 48GB GPUs.
    \item[] Guidelines:
    \begin{itemize}
        \item The answer \answerNA{} means that the paper does not include experiments.
        \item The paper should indicate the type of compute workers CPU or GPU, internal cluster, or cloud provider, including relevant memory and storage.
        \item The paper should provide the amount of compute required for each of the individual experimental runs as well as estimate the total compute. 
        \item The paper should disclose whether the full research project required more compute than the experiments reported in the paper (e.g., preliminary or failed experiments that didn't make it into the paper). 
    \end{itemize}
    
\item {\bf Code of ethics}
    \item[] Question: Does the research conducted in the paper conform, in every respect, with the NeurIPS Code of Ethics \url{https://neurips.cc/public/EthicsGuidelines}?
    \item[] Answer: \answerYes{} 
    \item[] Justification: The work follows the NeurIPS Code of Ethics.
    \item[] Guidelines:
    \begin{itemize}
        \item The answer \answerNA{} means that the authors have not reviewed the NeurIPS Code of Ethics.
        \item If the authors answer \answerNo, they should explain the special circumstances that require a deviation from the Code of Ethics.
        \item The authors should make sure to preserve anonymity (e.g., if there is a special consideration due to laws or regulations in their jurisdiction).
    \end{itemize}

\item {\bf Broader impacts}
    \item[] Question: Does the paper discuss both potential positive societal impacts and negative societal impacts of the work performed?
    \item[] Answer: \answerNA{} 
    \item[] Justification: The paper focuses on a technical compression method for feed-forward 3D Gaussian representations and does not introduce a deployment system or application-specific decision-making pipeline. We do not identify direct societal impacts.
    \item[] Guidelines:
    \begin{itemize}
        \item The answer \answerNA{} means that there is no societal impact of the work performed.
        \item If the authors answer \answerNA{} or \answerNo, they should explain why their work has no societal impact or why the paper does not address societal impact.
        \item Examples of negative societal impacts include potential malicious or unintended uses (e.g., disinformation, generating fake profiles, surveillance), fairness considerations (e.g., deployment of technologies that could make decisions that unfairly impact specific groups), privacy considerations, and security considerations.
        \item The conference expects that many papers will be foundational research and not tied to particular applications, let alone deployments. However, if there is a direct path to any negative applications, the authors should point it out. For example, it is legitimate to point out that an improvement in the quality of generative models could be used to generate Deepfakes for disinformation. On the other hand, it is not needed to point out that a generic algorithm for optimizing neural networks could enable people to train models that generate Deepfakes faster.
        \item The authors should consider possible harms that could arise when the technology is being used as intended and functioning correctly, harms that could arise when the technology is being used as intended but gives incorrect results, and harms following from (intentional or unintentional) misuse of the technology.
        \item If there are negative societal impacts, the authors could also discuss possible mitigation strategies (e.g., gated release of models, providing defenses in addition to attacks, mechanisms for monitoring misuse, mechanisms to monitor how a system learns from feedback over time, improving the efficiency and accessibility of ML).
    \end{itemize}
    
\item {\bf Safeguards}
    \item[] Question: Does the paper describe safeguards that have been put in place for responsible release of data or models that have a high risk for misuse (e.g., pre-trained language models, image generators, or scraped datasets)?
    \item[] Answer: \answerNA{} 
    \item[] Justification: The paper poses no such risks.
    \item[] Guidelines:
    \begin{itemize}
        \item The answer \answerNA{} means that the paper poses no such risks.
        \item Released models that have a high risk for misuse or dual-use should be released with necessary safeguards to allow for controlled use of the model, for example by requiring that users adhere to usage guidelines or restrictions to access the model or implementing safety filters. 
        \item Datasets that have been scraped from the Internet could pose safety risks. The authors should describe how they avoided releasing unsafe images.
        \item We recognize that providing effective safeguards is challenging, and many papers do not require this, but we encourage authors to take this into account and make a best faith effort.
    \end{itemize}

\item {\bf Licenses for existing assets}
    \item[] Question: Are the creators or original owners of assets (e.g., code, data, models), used in the paper, properly credited and are the license and terms of use explicitly mentioned and properly respected?
    \item[] Answer: \answerYes{} 
    \item[] Justification: The paper cites the datasets, pretrained FF3DGS models, and compression baselines used in the experiments.
    \item[] Guidelines:
    \begin{itemize}
        \item The answer \answerNA{} means that the paper does not use existing assets.
        \item The authors should cite the original paper that produced the code package or dataset.
        \item The authors should state which version of the asset is used and, if possible, include a URL.
        \item The name of the license (e.g., CC-BY 4.0) should be included for each asset.
        \item For scraped data from a particular source (e.g., website), the copyright and terms of service of that source should be provided.
        \item If assets are released, the license, copyright information, and terms of use in the package should be provided. For popular datasets, \url{paperswithcode.com/datasets} has curated licenses for some datasets. Their licensing guide can help determine the license of a dataset.
        \item For existing datasets that are re-packaged, both the original license and the license of the derived asset (if it has changed) should be provided.
        \item If this information is not available online, the authors are encouraged to reach out to the asset's creators.
    \end{itemize}

\item {\bf New assets}
    \item[] Question: Are new assets introduced in the paper well documented and is the documentation provided alongside the assets?
    \item[] Answer: \answerYes{} 
    \item[] Justification: We release code and pretrained models with documentation covering installation, dataset preparation, training, and evaluation.
    \item[] Guidelines:
    \begin{itemize}
        \item The answer \answerNA{} means that the paper does not release new assets.
        \item Researchers should communicate the details of the dataset\slash code\slash model as part of their submissions via structured templates. This includes details about training, license, limitations, etc. 
        \item The paper should discuss whether and how consent was obtained from people whose asset is used.
        \item At submission time, remember to anonymize your assets (if applicable). You can either create an anonymized URL or include an anonymized zip file.
    \end{itemize}

\item {\bf Crowdsourcing and research with human subjects}
    \item[] Question: For crowdsourcing experiments and research with human subjects, does the paper include the full text of instructions given to participants and screenshots, if applicable, as well as details about compensation (if any)? 
    \item[] Answer: \answerNA{} 
    \item[] Justification: The paper does not involve crowdsourcing nor research with human subjects.
    \item[] Guidelines:
    \begin{itemize}
        \item The answer \answerNA{} means that the paper does not involve crowdsourcing nor research with human subjects.
        \item Including this information in the supplemental material is fine, but if the main contribution of the paper involves human subjects, then as much detail as possible should be included in the main paper. 
        \item According to the NeurIPS Code of Ethics, workers involved in data collection, curation, or other labor should be paid at least the minimum wage in the country of the data collector. 
    \end{itemize}

\item {\bf Institutional review board (IRB) approvals or equivalent for research with human subjects}
    \item[] Question: Does the paper describe potential risks incurred by study participants, whether such risks were disclosed to the subjects, and whether Institutional Review Board (IRB) approvals (or an equivalent approval/review based on the requirements of your country or institution) were obtained?
    \item[] Answer: \answerNA{} 
    \item[] Justification: The paper does not involve crowdsourcing nor research with human subjects.
    \item[] Guidelines:
    \begin{itemize}
        \item The answer \answerNA{} means that the paper does not involve crowdsourcing nor research with human subjects.
        \item Depending on the country in which research is conducted, IRB approval (or equivalent) may be required for any human subjects research. If you obtained IRB approval, you should clearly state this in the paper. 
        \item We recognize that the procedures for this may vary significantly between institutions and locations, and we expect authors to adhere to the NeurIPS Code of Ethics and the guidelines for their institution. 
        \item For initial submissions, do not include any information that would break anonymity (if applicable), such as the institution conducting the review.
    \end{itemize}

\item {\bf Declaration of LLM usage}
    \item[] Question: Does the paper describe the usage of LLMs if it is an important, original, or non-standard component of the core methods in this research? Note that if the LLM is used only for writing, editing, or formatting purposes and does \emph{not} impact the core methodology, scientific rigor, or originality of the research, declaration is not required.
    \item[] Answer: \answerNA{} 
    \item[] Justification: The core method development in this research does not involve LLMs as any important, original, or non-standard components.
    \item[] Guidelines:
    \begin{itemize}
        \item The answer \answerNA{} means that the core method development in this research does not involve LLMs as any important, original, or non-standard components.
        \item Please refer to our LLM policy in the NeurIPS handbook for what should or should not be described.
    \end{itemize}

\end{enumerate}